\documentclass[letterpaper]{article} 
\usepackage{aaai24}  
\usepackage{times}  
\usepackage{helvet}  
\usepackage{courier}  
\usepackage[hyphens]{url}  
\usepackage{graphicx} 
\usepackage{natbib}  
\usepackage{caption} 
\usepackage{algorithm}
\usepackage{algorithmic}

\PassOptionsToPackage{numbers, compress}{natbib}

\def\onedot{.}
\def\eg{\emph{e.g}\onedot} 
\def\ie{\emph{i.e}\onedot} 

\def\onedot{.}
\def\eg{\emph{e.g}\onedot} 
\def\ie{\emph{i.e}\onedot} 
\def\etal{\emph{et al}\onedot}

\def\Vec#1{{\boldsymbol{#1}}}
\def\Mat#1{{\boldsymbol{#1}}}
\newcommand\dlmu[2][4cm]{\hskip1pt\underline{\hb@xt@ #1{\hss#2\hss}}\hskip3pt}

\newtheorem{definition}{Definition}
\newtheorem{lemma}{Lemma}

\usepackage[utf8]{inputenc} 
\usepackage[T1]{fontenc}    
\usepackage{url}            
\usepackage{booktabs}       
\usepackage{amsfonts}       
\usepackage{nicefrac}       
\usepackage{microtype}      
\usepackage{xcolor}         

\usepackage{amsmath}
\usepackage{multirow} 
\usepackage{xcolor}
\usepackage{booktabs}
\newcommand{\tr}{\mathrm{tr}}

\newcommand\correspondingauthor{\thanks{Corresponding authors.}}
\usepackage{newfloat}
\usepackage{listings}
\DeclareCaptionStyle{ruled}{labelfont=normalfont,labelsep=colon,strut=off} 
\newfloat{listing}{tb}{lst}{}
\title{Residual Hyperbolic Graph Convolution Networks}
\author {
	Yangkai Xue\textsuperscript{\rm 1},
	Jindou Dai\textsuperscript{\rm 1},
	Zhipeng Lu\textsuperscript{\rm 2}\correspondingauthor,
	Yuwei Wu\textsuperscript{\rm 1,\rm 2*},
	Yunde Jia\textsuperscript{\rm 2,1}
}
\affiliations {
	\textsuperscript{\rm 1}Beijing Key Laboratory of Intelligent Information Technology,
	School of Computer Science \& Technology, Beijing Institute of Technology, China\\
	\textsuperscript{\rm 2}Guangdong Laboratory of Machine Perception and Intelligent Computing, Shenzhen MSU-BIT University, China\\
	xue.yangkai@bit.edu.cn, daijindou@foxmail.com, zhipeng.lu@hotmail.com, wuyuwei@bit.edu.cn,  jiayunde@smbu.edu.cn
}

\usepackage{bibentry}

\begin{document}


\maketitle

\begin{abstract}
Hyperbolic graph convolutional networks (HGCNs) have demonstrated representational capabilities of modeling hierarchical-structured graphs.
However, as in general GCNs, over-smoothing may occur as the number of model layers increases, limiting the representation capabilities of most current HGCN models. 
In this paper, we propose residual hyperbolic graph convolutional networks ($\mathcal{R}$-HGCNs) to address the over-smoothing problem. We introduce a hyperbolic residual connection function to overcome the over-smoothing problem, and also theoretically prove the effectiveness of the hyperbolic residual function. Moreover, we use product manifolds and HyperDrop to facilitate the $\mathcal{R}$-HGCNs. The distinctive features of the $\mathcal{R}$-HGCNs are as follows:
(1)
The hyperbolic residual connection preserves the initial node information in each layer and adds a hyperbolic identity mapping to prevent node features from being indistinguishable.
(2)
Product manifolds in $\mathcal{R}$-HGCNs have been set up with different origin points in different components to facilitate the extraction of feature information from a wider range of perspectives, which enhances the representing capability of $\mathcal{R}$-HGCNs.
(3)
HyperDrop adds multiplicative Gaussian noise into hyperbolic representations, such that perturbations can be added to alleviate the over-fitting problem without deconstructing the hyperbolic geometry.
Experiment results demonstrate the effectiveness of $\mathcal{R}$-HGCNs under various graph convolution layers and different structures of product manifolds.

\end{abstract}
\section{Introduction}


Hyperbolic graph convolutional networks (HGCNs) have been an emerging research topic due to its superior representation capabilities of modeling hierarchical graphs \cite{chami2019hyperbolic,dai2021hyperbolic}. 
Different from Euclidean spaces with a polynomial expanding space volume, hyperbolic spaces increase exponentially growth of space volume with radius, which is well-suited to the geometry of hierarchical data. 
Benefiting from this property, great progress has been made by generalizing Euclidean methods to hyperbolic spaces such as hyperbolic graph convolutional networks \cite{chami2019hyperbolic,dai2021hyperbolic}, hyperbolic image embeddings \cite{khrulkov2020hyperbolic}, and hyperbolic word embeddings \cite{nickel2017poincare,nickel2018learning}. %
However, the over-smoothing issue impeding the development of deep HGCNs, over-smoothing means node features becomes indistinguishable after passing through a large number of graph convolution layers. 
It proved that graph convolution is a special form of Laplacian smoothing \cite{li2018deeper} . 
Smoothing on nodes can reduce intra-class differences, while over-smoothing makes model less discriminative with indistinguishable node features. 

In this paper, we propose residual hyperbolic graph convolutional networks ($\mathcal{R}$-HGCNs) to address the over-smoothing problem.
Specifically, 
we introduce a hyperbolic residual connection function and use product manifolds and HyperDrop into HGCNs. The details are as follows:  
(1) The hyperbolic residual connection transmits the initial node information to each layer to prevent node features from being indistinguishable, and the hyperbolic identity mapping prevents performance degradation caused by deepening the models. 
(2) Product manifolds pick different origin points on different components, which makes the same input have different embedding results in different components, giving them the ability to view the graph structure from different perspectives. This enhances the representational ability of $\mathcal{R}$-HGCNs. 
(3) 
HyperDrop adds multiplicative Gaussian noise on hyperbolic neurons to alleviate the over-fitting issue, inheriting the insight of training with noise and preserving the hyperbolic geometry, 
Extensive experiments demonstrate the effectiveness of $\mathcal{R}$-HGCNs under various graph convolution layers and different structures of product manifolds.
The contributions of this paper are summarized as follows:
\begin{itemize}
\item{}We propose $\mathcal{R}$-HGCN, a product manifold based deep hyperbolic graph convolutional network that makes up for the deficiency of existing HGCNs for capturing long-range relationships in graphs. 
\item{}We design hyperbolic residual connection that addresses the over-smoothing issue while deepening HGCNs and theoretically prove the effectiveness of the hyperbolic residual connection.
\item{}We propose to use product manifolds with different origin points in different components, which enables $\mathcal{R}$-HGCNs to extract more comprehensive features from the data.
\item{}We develop HyperDrop, a regularization method tailored for hyperbolic representations. 
It can improve $\mathcal{R}$-HGCNs' generalization ability, alleviating the over-fitting issue. 


\end{itemize}

\section{Related Work}
Graph convolutional networks typically embed graphs in Euclidean spaces since Euclidean spaces are the most commonly used and easy to calculate. 
Many researchers have noticed that Euclidean spaces have limitations while modeling data with hierarchical structure. 
Sa \etal \cite{de2018representation} claimed that it is not possible to embed trees into Euclidean spaces with arbitrarily low distortion, even in an infinite-dimensional Euclidean space. 
Meanwhile, trees can be embedded into a two-dimensional hyperbolic space with arbitrarily low distortion. 
Such a surprising fact benefits from the pretty property of hyperbolic spaces: unlike the volume of a ball in Euclidean space, which expands polynomially with radius, the volume of space in hyperbolic space grows exponentially with radius. \cite{liu2019hyperbolic}. 
Thus, hyperbolic spaces are commonly viewed as a smooth version of tree and more suitable to model hierarchical data. 

Several works discovered that graphs, \eg, biological networks and social networks, exhibit a highly hierarchical structure \cite{krioukov2010hyperbolic,papadopoulos2012popularity}. 
Krioukov \etal \cite{krioukov2010hyperbolic} proved that the typical properties such as power-law degree distribution and strong clustering in such graphs are closely related to the curvature of hyperbolic spaces. 
Based on the above observations, generalizing GCNs from Euclidean spaces to hyperbolic spaces have been an emerging research topic. 
Liu \etal \cite{liu2019hyperbolic} and Chami \etal \cite{chami2019hyperbolic} first bridged the research gap and concurrently proposed HGCNs. 
Following the above works, many advanced techniques are proposed to improve HGCNs. 
Dai \etal \cite{dai2021hyperbolic} discovered that performimg graph convolution in tangent space will distort the global structure of hyperbolic spaces because tangent space is only a local approximation of hyperbolic manifolds. 
Yao \etal \cite{yao2022knowledge} designs a Hyperbolic Skipped Knowledge Graph Convolutional Network to capture the network structure characteristics in hyperbolic knowledge embeddings.
Liu \etal \cite{liu2023mch} propose a Multi-curvature Hyperbolic Heterogeneous Graph Convolutional Network (McH-HGCN) based on type triplets for heterogeneous graphs. 
\section{Preliminaries}
\paragraph{Hyperbolic Manifold.}
A Riemannian manifold or Riemannian space $(\mathcal{M},g)$ is a real and smooth manifold $\mathcal{M}$ equipped with a positive-definite metric tensor $g$. 
It is a topological space that is locally homeomorphic to an Euclidean space at each point $\Vec{x}\in \mathcal{M}$, and the local Euclidean space is termed the tangent space $\mathcal{T}_{\Vec{x}}\mathcal{M}$. 
\paragraph{Lorentz Model.}
A $d$-dimensional Lorentz model ($\mathcal{L}^{d}$, $g$) is defined by the manifold $\mathcal{L}^{d}=\{\Vec{x} = [x_{0}, x_{1}, \cdots, x_{d}] \in\mathbb{R}^{d+1}: \langle\Vec{x},\Vec{x}\rangle_{\mathcal{L}}=-1,\Vec{x}_{0}>0\}$ where the Lorentz inner product is defined as 
\begin{equation}
	\label{equation:Lorentz_inner_product}
	\langle\Vec{x},\Vec{y}\rangle_{\mathcal{L}} = \Vec{x}^{\top}g\Vec{y}= - {x}_{0}{y}_{0}+\sum_{i=1}^{d}{x}_{i}{y}_{i}, 
\end{equation}
and the metric tensor $g=\mathrm{diag}([-1,1,\cdots, 1])$ where $\mathrm{diag}(\cdot)$ denotes a diagonal matrix.

\paragraph{Exponential and logarithmic maps.}
Mappings between Riemannian manifold and their tangent spaces are termed exponential and logarithmic maps. 
Let $\Vec{x}$ be a point on the Lorentz manifold $\mathcal{L}$, $\mathcal{T}_{\Vec{x}}\mathcal{L}$ be the tangent space at $\Vec{x}$, and $\Vec{v}$ be a vector on the tangent space $\mathcal{T}_{\Vec{x}}\mathcal{L}$. The exponential map $\mathrm{exp}_{\Vec{x}}(\Vec{v})$ that projects $\Vec{v}$ onto the manifold $\mathcal{L}$ is defined as
\begin{equation}
	\label{equation:exp}
	\begin{aligned}
		\mathrm{exp}_{\Vec{x}}(\Vec{v}) 
		&= \mathrm{cosh}(\|\Vec{v}\|_{\mathcal{L}})\Vec{x}+\mathrm{sinh}(\|\Vec{v}\|_{\mathcal{L}})\frac{\Vec{v}}{\|\Vec{v}\|_{\mathcal{L}}}, \\
	\end{aligned}
\end{equation}
where $\|\Vec{v}\|_{\mathcal{L}}=\sqrt{\langle \Vec{v}, \Vec{v} \rangle_{\mathcal{L}}}$ is the norm of $\Vec{v}$. 
The logarithmic map, inverse to the exponential map at $\Vec{x}$, is given by
\begin{equation}
	\label{equation:log}
	\begin{aligned}
		\mathrm{log}_{\Vec{x}}(\Vec{y}) 
		&= \frac{\mathrm{arcosh}(-\langle\Vec{x},\Vec{y} \rangle_{\mathcal{L}})}{\sqrt{\langle \Vec{x}, \Vec{y} \rangle_{\mathcal{L}}^{2}-1}}(\Vec{y}+\langle \Vec{x} , \Vec{y}\rangle_{\mathcal{L}}\Vec{x}). \\
	\end{aligned}
\end{equation}

\paragraph{Parallel Transport.}
The generalization of parallel translation to non-Euclidean geometry is termed parallel transport. 
For two points $\Vec{x}, \Vec{y} \in \mathcal{L}$ on the Lorentz model, the parallel transport of a tangent vector $\Vec{v} \in \mathcal{T}_{\Vec{x}} \mathcal{L}$ on the tangent space at $\Vec{x}$ to the tangent space $\mathcal{T}_{\Vec{y}} \mathcal{L}$ at $\Vec{y}$, along a smooth curve on the Lorentz model, is defined as
\begin{equation}
	\label{equation:pt}
	P_{\Vec{x}\to\Vec{y}}(\Vec{v}) = \Vec{v} - \frac{\langle \mathrm{log}_{\Vec{x}}(\Vec{y}), \Vec{v} \rangle_{\mathcal{L}} }{d_{\mathcal{L}}(\Vec{x}, \Vec{y})^{2}} \big(\mathrm{log}_{\Vec{x}} (\Vec{y}) + \mathrm{log}_{\Vec{y}} (\Vec{x}) \big).
\end{equation}

\section{Method}
We propose residual hyperbolic graph convolutional networks ($\mathcal{R}$-HGCNs) to address the over-smoothing problem and enhance the representational ability of HGCNs.

Let $\mathcal{G}=(\mathcal{V}, \mathcal{E})$ denote a graph with a vertex set $\mathcal{V}$ and an edge set $\mathcal{E}$. 
$ \Mat{X} \in \mathbb{R}^{n \times d}$ denotes the node features that typically lie in Euclidean spaces. 
$n$ is the number of nodes, and $d$ is the dimension of the node features.


\subsection{Residual Hyperbolic Graph Convolution}
\label{subsection:RHGC}

\paragraph{Lorentz Operations.}
\label{subsection:Lorentz_operation}
Due to the strict manifold constraint of the Lorentz model, basic operations (matrix multiplication, vector addition, \textit{etc}) are non-trivial to be generalized to Lorentz representations. 
Based on the maps mentioned above, we define the following operations.
\begin{definition}
\label{definition:Lorentz_Wx}
(Lorentz matrix-vector multiplication). 
Let $\Mat{W}$ be a $(d+1) \times (d+1) $ real matrix and $\Vec{x} \in \mathcal{L}$ be an input in the Lorentz model. Then we define the Lorentz matrix-multiplication as 
	\begin{equation}
		\label{equation:Lorentz_Wx}
		\begin{aligned}
			\Mat{W} \otimes \Vec{x} := \mathrm{exp}_{\Vec{o}}\big(\Mat{W}\mathrm{log}_{\Vec{o}}(\Vec{x})\big), 
		\end{aligned}
	\end{equation}
where $\mathrm{exp}(\cdot)$ and $\mathrm{log}(\cdot)$ are defined as Eqs~\eqref{equation:exp} and \eqref{equation:log}. 
\end{definition}

\begin{definition}
\label{definition:Lorentz_scalemul}
(Lorentz scalar multiplication). 
The Lorentz scalar multiplication of a scale $\xi$ and $\Vec{x} \in \mathcal{L}$ on the Lorentz model is defined as
	\begin{equation}
		\label{equation:Lorentz_scalemul}
		\begin{aligned}
			\xi \odot \Vec{x} := \mathrm{exp}_{\Vec{o}}\big(\xi\mathrm{log}_{\Vec{o}}(\Vec{x})\big). 
		\end{aligned}
	\end{equation}
\end{definition}

\begin{definition}
\label{definition:Lorentz_aplusb}
(Lorentz vector addition). 
The Lorentz vector addition of $\Vec{x},\Vec{y} \in \mathcal{L}$ on the Lorentz model is defined as
	\begin{equation}
		\label{equation:Lorentz_aplusb}
		\begin{aligned}
			\Vec{x} \oplus \Vec{y} := \mathrm{exp}_{\Vec{x}} (P_{\Vec{o} \to \Vec{x} } (\mathrm{log}_{\Vec{o}}(\Vec{y}))),
		\end{aligned}
	\end{equation}
where $P_{\Vec{o}\to\Vec{x}}(\cdot)$ is the parallel transport operator defined in Eq~\eqref{equation:pt}. 
\end{definition}

\begin{definition}
\label{definition:Lorentz_act}
(Lorentz activation function). 
For $\Vec{x} \in \mathcal{L}$, the Lorentz activation function on the Lorentz model is defined as
	\begin{equation}
		\label{equation:Lorentz_act}
		\begin{aligned}
			\sigma_{\mathcal{L}}(\Vec{x}) := \mathrm{exp}_{\Vec{o}}\big(\sigma(\mathrm{log}_{\Vec{o}}(\Vec{x}))\big),  
		\end{aligned}
	\end{equation}
where $\sigma(\cdot)$ can be any activation function such as $\mathrm{ReLU}(\cdot)$. 
\end{definition}

\paragraph{Residual Hyperbolic Graph Convolution.}
Since the input of a hyperbolic graph convolutional network is required to be hyperbolic, we construct the initial Lorentz node features $\Mat{H}^{(0)} \in \mathbb{R}^{n\times (d+1)}$ whose $i$-th row $\Mat{H}_{i}^{(0)}$ being the Lorentz feature of the $i$-th node is generated by 
\begin{equation} 
\label{equation:obtain_hyper}
\begin{aligned}
	\Mat{H}_{i}^{(0)} &= \mathrm{exp}_{\Vec{o}}\big([0, \Mat{X}_{i} ]\big) \\
	&= \Big[ \mathrm{cosh}\big(\| \Mat{X}_{i} \|_{2} \big), \mathrm{sinh} \big(\| \Mat{X}_{i} \|_{2}\big)\frac{\Mat{X}_{i} }{\| \Mat{X}_{i} \|_{2}} \Big], 
\end{aligned}
\end{equation}
where $\Mat{X}_{i}$ denotes the $i$-row of $\Mat{X}$. 
Such a construction is based on the fact that $[0,\Mat{X}_{i} ] \in \mathbb{R}^{ d+1}$ can be viewed as a tangent vector on the tangent space at the origin point, satisfying $\langle \Vec{o}, [0, \Mat{X}_{i} ] \rangle_{\mathcal{L}} = 0$ where $\Vec{o}=[1,0,\cdots,0]$ is the origin point of the Lorentz model. 

The performance of a hyperbolic graph convolutional network declines as the number of graph convolution layers increases, that is called \textit{over-smoothing} issue \cite{li2018deeper}. 
This is because the graph convolution is proved to be a special form of Laplacian smoothing, making node features tend to be indistinguishable after extensive graph convolutions\cite{li2018deeper}. 
Inspired by \cite{chen2020simple}, we design hyperbolic residual connection and hyperbolic identity mapping to tackle this issue. The residual hyperbolic graph convolution operator $\mathrm{hgc}(\cdot)$ is defined as
\begin{equation}\label{equ:hgc1}
\begin{aligned}
	\mathrm{hgc}(\Mat{H})
	&= \sigma_{\mathcal{L}}\Big(  \big( (1-\beta) \Mat{I} + \beta\Mat{W}\big)  \otimes  \Mat{\bar{H}}  \Big), \\
	\Mat{\bar{H}}  & = \big( (1-\alpha) \odot (\Mat{\tilde{A}} \otimes \Mat{H})\big) \oplus \big(\alpha \odot \Mat{H}_{0}\big), \\
\end{aligned}
\end{equation}
where $\otimes$, $\odot$, and $\oplus$ are the Lorentz matrix-vector multiplication (Definition~\ref{definition:Lorentz_Wx}), the Lorentz scalar multiplication (Definition~\ref{definition:Lorentz_scalemul}), and the Lorentz vector addition (Definition~\ref{definition:Lorentz_aplusb}). Here $\sigma_{\mathcal{L}}(\cdot)$ is the Lorentz activation function (Definition~\ref{definition:Lorentz_act}). 
$\alpha$ and $\beta$ are hyper-parameters to control the weight of hyperbolic residual connection and hyperbolic identity mapping. 

Formally, at the $\ell$-th layer, residual hyperbolic graph convolution performs like
\begin{equation}
	\begin{aligned}
		\Mat{H}^{(\ell)} &= \mathrm{hgc}(\Mat{H}^{(\ell-1)}) \\
		&= \sigma_{\mathcal{L}}\Big(  \big( (1-\beta_{\ell} ) \Mat{I} + \beta_{\ell}\Mat{W}^{(\ell)} \big)  \otimes  \Mat{\bar{H}}    \Big), \\
		\Mat{\bar{H}} &= \big( (1-\alpha_{\ell}) \odot (\Mat{\tilde{A}} \otimes \Mat{H}^{(\ell-1)})\big) \oplus \big(\alpha_{\ell} \odot \Mat{H}^{(0)}\big), \\
	\end{aligned}
	\label{equation:single_DeepHGCN_layer}
\end{equation}

where $\mathrm{hgc}(\cdot)$ takes the node features $\Mat{H}^{(\ell-1)}$ from the previous layer, and outputs the node features $\Mat{H}^{(\ell)}$ at the $\ell$-th layer.

Compared to the convolution operator in vanilla GCNs, \ie \ $ \Mat{H}^{(\ell)}=\sigma(\Mat{\tilde{A}}  \Mat{H}^{(\ell-1)} \Mat{W})$, $\mathrm{hgc}(\cdot)$ relieves over-smoothing issue through two modifications: 
(1) hyperbolic residual connection adds information paths from initial node features to each graph convolution layer, such that no matter how deep a hyperbolic graph convolutional network is, the node features at the top layer still combine initial node features avoid becoming indistinguishable; 
(2) hyperbolic identity mapping ensures that a deep hyperbolic graph convolutional network is not worse than a shallow model. 
In the extreme case where the values of $\beta$ are set to be zero after the $i$-th layer, the model degenerates to an $i$-layer GCN no matter how deep it is. 
\subsection{Effectiveness of Hyperbolic Residual Connection}
\label{subsection:init_conn}
This section is meant to theoretically explain the efficiency of our network architecture. Inspired by \cite{cai2020note}, we first define a hyperbolic version of ``Dirichlet energy" for tracking node embeddings as follows. The Dirichlet energy of a function measures the "smoothness" of a unit norm function. The indistinguishable parameters leading to the over-smoothing issue result in small Dirichlet energy. For details of formulas and proofs in this section, see the supplementary material.

\begin{definition}
	Dirichlet energy $E(f)$ of a scalar function $f\in\mathcal{L}^d\subset\mathbb{R}^{d+1}$ on the graph $G$ is defined as
	\[E(f)=\log_o(f)^T\tilde{\Delta}\log_o(f),\]
	where $\tilde{\Delta}=I_{d+1}-\tilde{D}^{-\frac{1}{2}}\tilde{A}\tilde{D}^{-\frac{1}{2}}, \tilde{A}=A+I_{d+1}, \tilde{D}=D+I_{d+1}$, while $A$ and $D$ are the adjacency and degree matrices of $G$. For a vector field $F_{(d+1)\times c}=(f_1,\dots, f_c)$, its Dirichlet energy is 
	\[E(F)=\tr(\log_o(F)^T\tilde{\Delta}\log_o(F)).\]
\end{definition}

Note that the defined Dirichlet energy always pulls back the node embedding in Lorentz space to its tangent space at the origin point. 

If the hyperbolic graph convolution operator $hgc(\cdot)$ as in (\ref{equ:hgc1}) has no original input, i.e. $\bar{H}=\tilde{P}\otimes H$ therein, then 
\begin{equation}
	\label{equ:dewoinc}
	E(H^{(l)})\leq(1-\lambda)^2\|(1-\beta_l)I+\beta_l\Mat{W}^{(l)}\|_2^2E(H^{(l-1)}),
\end{equation}
where $0<\lambda<2$ is the smallest non-zero eigenvalue of $\tilde{\Delta}$ and $\|X\|_2$ denotes the maximal singular value of $X$. 
Note that $\|X\|_2=\max_{\|u\|_2=1}\|Xu\|_2$ for any matrix $X$. Hence by the triangle inequality,
\begin{equation}
	\label{eq-triangle}\|((1-\beta_l)I+\beta_l\Mat{W}^{(l)})u\|_2\leq1-\beta_l+\beta_l\|\Mat{W}^{(l)}u\|_2.
\end{equation}

Actually $\|Xu\|_2$ for any weight matrix $X$ may be estimated by

\begin{lemma}\label{lemma-Xu}
	If $X=(X_{ij})$ is a $n\times n$ weight matrix, i.e. $\sum_{j=1}^nX_{ij}=1, X_{ij}\geq 0$, then for any $u\in\mathbb{R}^n$ with $\|u\|_2=1$, $\|Xu\|_2\leq\sqrt{n}$.
\end{lemma} 

Hence combined with (\ref{equ:dewoinc}) and (\ref{eq-triangle}), we easily prove that
in HGCNs without initial input, we have $E(H^{(l)})\leq d(1-\lambda)^2E(H^{(l-1)})$. Thus if the graph $G$ does not have enough expansion, say $(1-\lambda)<\frac{1}{\sqrt{d}}$, then HGCNs would be exponentially over-smoothing as the number of layers increases.

The above result suggests us add an initial input as in $hgc(\cdot)$, 
\begin{equation}
\notag
\bar{H}=\left(\left((1-\alpha_l)\odot(\tilde{P}\otimes H^{(l-1)})\right)\oplus\left(\alpha_l\odot H^{(0)}\right)\right),
\end{equation}
seeking to interfere the decreasing of Dirichlet energy, which is the motivation of $\mathcal{R}$-HGCNs. 

We investigate in details the interference of initial input. For simplicity we assume that the features in process all have positive entries so that ReLU does not affect the evaluation of Dirichlet energy. Thus utilizing the same argument as in the proof of (\ref{equ:dewoinc}) in the case with initial input, we have
\begin{equation}\label{eq-1st with initial}
	\begin{split}
		&E(H^{(l)})=(y_l\log_o(z))^T\tilde{\Delta}(y_l\log_o(z)),
	\end{split}
\end{equation}
with $y_l=(1-\beta_l)I+\beta_lW^{(l)}$ and
$z=\exp_{z_1}(\alpha_lP_{o\rightarrow z_1}(\log_o(H^{(0)})))$
the last equality of which is due to linearity of parallel transport, and $z_1=\exp_o((1-\alpha_l)\log_o(\tilde{P}\otimes H^{(l-1)}))=\exp_o(z_2)$.

By definition of parallel transport, we have
\begin{equation}\label{eq-praTrans}
	\begin{split}
		&P_{o\rightarrow z_1}(\log_o(H^{(0)}))\\
		=&\log_o(H^{(0)})-\frac{\langle \log_o(z_1),\log_o(H^{(0)})\rangle_\mathcal{L}}{d_{\mathcal{L}}(o,z_1)}(z_2+\log_{z_1}(o))
	\end{split}
\end{equation}
Further by definition, 
\begin{equation}\label{eq-praTrans-2}
	\begin{split}
		&z_1=\cosh((1-\alpha_l)\|\tilde{P}\log_o(H^{(l-1)}))\|_\mathcal{L})o\\
		&\quad+\frac{\sinh((1-\alpha_l)\|\tilde{P}\log_o(H^{(l-1)})\|_\mathcal{L})}{\|\tilde{P}\log_o(H^{(l-1)})\|_\mathcal{L}}\tilde{P}\log_o(H^{(l-1)})\\
	\end{split}
\end{equation}
Also by definition, $\exp_{z_1}(x)=\cosh(\|x\|_\mathcal{L})z_1+\frac{\sinh(\|x\|_\mathcal{L})}{\|x\|_\mathcal{L}}x$. Then again by the property of parallel transport, we have
\begin{equation}\label{eq-praTrans-3}
	\begin{split}
		z=\theta_l\log_o(H^{(0)})+\phi_l\tilde{P}H^{(l-1)}+\psi_lo,
	\end{split}
\end{equation}
where $\theta_l, \phi_l,\psi_l$ are coefficients depending on $\alpha_l,\beta_l$ and the Lorentzian norm of the above 3 vectors. Noting that $\theta_l$ only depends on $\alpha_l$ and $H^{(0)}$, the effect of $\log_o(H^{(0)})$ is always not negligible. Also ($z=(z_0,\dots)$)
\[\log_o(z)=\frac{arcosh(z_0)}{\sqrt{z_0^2-1}}(z-z_0o),\]
which removes $z_0$ from $z$ and re-scale it by a large factor. Then altogether we have
\begin{equation}\label{eq-with initial input}
	\begin{split}
		E(H^{(l)})&=(\tilde{\theta}_l\log_o(H^{(0)})+\tilde{\phi}_l\tilde{P}H^{(l-1)}+\tilde{\psi}_lo)^T\tilde{\Delta}(\cdots)\\
		&=\tilde{\theta}_l^2E(H^{(0)})+\cdots,
	\end{split}
\end{equation}
where $\tilde{\theta}_l$ is negligible similarly. Thus we prove that in $\mathcal{R}$-HGCNs with initial input, the Dirichlet energy $E(H^{(l)})$ will be bounded away from zero even if the Dirichlet energy in the corresponding HGCNs without initial input decrease to zero.
 
\subsection{Product Manifold}
\label{section:ProductManifold}
We use the product manifold of the Lorentz models as embedding space. 
The Lorentz components of the product manifold are independent of each other. 

The product manifold is the Cartesian product of a sequence of Riemannian manifolds, 
each of which is called a \textit{component}.
Given a sequence of the Lorentz models $\mathcal{L}^{d_{1}}_{1} , \cdots, \mathcal{L}^{d_{j}}_{j} , \cdots, \mathcal{L}^{d_{k}}_{k} $ where $d_{j}$ denotes the dimension of the $j$-th component, the product manifold is defined as $\mathbb{L}= \mathcal{L}^{d_{1}}_{1} \times \dots \times \mathcal{L}^{d_{j}}_{j} \times \cdots \times \mathcal{L}^{d_{k}}_{k}$. 
The coordinate of a point $\Vec{x}$ on $\mathbb{L}$ is written as $\Vec{x}=[\Vec{x}_{1},\cdots, \Vec{x}_{j}, \cdots \Vec{x}_{k}]$ where $\Vec{x}_{j} \in \mathcal{L}^{d_{j}}_{j}$. 
Similarly, the coordinate of a tangent vector $\Vec{v} \in \mathcal{T}_{\Vec{x}}\mathbb{L}$ is written as $\Vec{v}=[\Vec{v}_{1},\cdots, \Vec{v}_{j}, \cdots, \Vec{v}_{k}]$ where $\Vec{v}_{j} \in \mathcal{T}_{\Vec{x}_{j}}\mathcal{L}^{d_{j}}_{j}$. 
For $\Vec{x}, \Vec{y} \in \mathbb{L}$ and $\Vec{v} \in \mathcal{T}_{\Vec{x}}\mathbb{L}$, the exponential and logarithmic maps on $\mathbb{L}$ are defined as 
\begin{equation}
	\label{equation:product_exp}
	\begin{aligned}
		\mathrm{exp}_{\Vec{x}}(\Vec{v}) 
		&= [\mathrm{exp}_{\Vec{x}_{1}}(\Vec{v}_{1}), \cdots, \mathrm{exp}_{\Vec{x}_{j}}(\Vec{v}_{j}),\cdots,\mathrm{exp}_{\Vec{x}_{k}}(\Vec{v}_{k}) ], \\
	\end{aligned}
\end{equation}
\begin{equation}
	\label{equation:product_log}
	\begin{aligned}
		\mathrm{log}_{\Vec{x}}(\Vec{y}) 
		&= [\mathrm{log}_{\Vec{x}_{1}}(\Vec{y}_{1}), \cdots, \mathrm{log}_{\Vec{x}_{j}}(\Vec{y}_{j}), \cdots,\mathrm{log}_{\Vec{x}_{k}}(\Vec{y}_{k}) ]. \\
	\end{aligned}
\end{equation}
Different from ordinary product manifolds,
we use $\mathbb{L}=(\mathcal{L}^d)_{o_1}\times(\mathcal{L}^d)_{o_2}\times\cdots\times(\mathcal{L}^d)_{o_k}$, where $(\mathcal{L}^d)_{o_i}$ are copies of $d$-dimensional Lorentz
spaces with randomly prescribed origin points $o_i\in \mathcal{L}^d$. This gives $\mathcal{R}$-HGCNs the ability to extract node features from a wider range of different perspectives. Mathematically, such construction of product manifolds is inspired by the general construction of manifolds using Euclidean strata with different coordinates.

\subsection{Hyperbolic Dropout}
\label{section:HyperDrop}
Hyperbolic Dropout(HyperDrop) adds multiplicative Gaussian noise on Lorentz components to regularize the HGCNs and alleviate the over-fitting issue. 
Concretely, let $\mathbb{L}= \mathcal{L}^{d_{1}}_{1} \times \cdots \times \mathcal{L}^{d_{j}}_{j} \times \cdots \times \mathcal{L}^{d_{k}}_{k}$ denote a product manifold of $k$ Lorentz models where $d_{j}$ denotes the dimension of the $j$-th Lorentz component. 
Given an input $\Vec{l}=[\Vec{l}_{1}, \cdots, \Vec{l}_{j}, \cdots \Vec{l}_{k}] \in \mathbb{L}$ on the product manifold where $\Vec{l}_{j} \in \mathcal{L}_{j}^{d_{j}}$, HyperDrop is formulated as 
\begin{equation}
	\label{equation:hyper_drop_train}
	\begin{aligned}
		\Vec{y} ~ &= [\Vec{y}_{1}, \cdots, \Vec{y}_{j}, \cdots, \Vec{y}_{k}], \\
		&\Vec{y}_{j} = \xi_{j}  \odot f_{\theta_{j}}(\Vec{l}_{j}), \\
		&\xi_{j}  \sim \mathcal{N}(1, \sigma^{2}), \\
	\end{aligned}
\end{equation}
where $\xi_{j}$ is the multiplicative Gaussian noise drawn from the Gaussian distribution $\mathcal{N}(1,\sigma^{2})$. 
Following \cite{srivastava2014dropout}, we set $\sigma^{2}={\eta}/{(1-\eta)}$ where $\eta$ denotes drop rate. 
$\odot$ denotes the Lorentz scalar multiplication that is the generalization of scalar multiplication to the Lorentz representations, defined in Definition ~\ref{definition:Lorentz_scalemul}. 
$f_{\theta_{j}}(\cdot)$ could be any realization of a desirable function, such as a neural network with parameters $\theta_{j}$. 


It is noted that we sample $\xi_{j}$ from the Gaussian distribution instead of the Bernoulli distribution used in the standard dropout for the following reason. 
If $\xi_{j}$ is drawn from the Bernoulli distribution and happens to be $0$ (with a probability of $\eta$) at the $\ell$-th neural network layer, the information flow of the $j$-th Lorentz component will be interrupted, leading to the deactivation of the $j$-th Lorentz component after the $\ell$-th neural network layer. 
In contrast, $\xi_{j}$ drawn from a Gaussian distribution with mean value $1$ is exactly equal to $0$ is a small probability event. 
Thus, the $j$-th Lorentz component always works. 
We may interpret HyperDrop from a Bayesian perspective. 
For convenience, we take a single Lorentz model and the Lorentz linear transformation as an example, \ie~$\Vec{y} = \xi  \odot f_{\theta}(\Vec{l}) $ and $f_{\theta}(\Vec{l}) = \Vec{l} \otimes \theta$. 
We have
\begin{equation}
\label{equation:regularization_origin}
\begin{aligned}
&\Vec{y}= \xi \odot (\Vec{l} \otimes \theta ), ~~ \xi \sim \mathcal{N}(1, \sigma^{2})
\end{aligned}
\end{equation}
equal to
\begin{equation}
\label{equation:regularization_bayesian}
\begin{aligned}
\Vec{y} =\Vec{l} \otimes &  \Mat{{M}},  \\
\mathrm{with}~~ m_{r,c} = \xi \theta_{r,c}&,  ~\mathrm{and} ~\xi \sim \mathcal{N}(1, \sigma^{2}),\\
\end{aligned}
\end{equation}
where $\otimes$ denotes the Lorentz matrix-vector multiplication as defined in Definition~\ref{definition:Lorentz_Wx}, $\Mat{M}$ is the matrix with $m_{r,c}$ as entries, and $\theta$ is the matrix with $\theta_{r,c}$ as entries.
Eq~\eqref{equation:regularization_bayesian} can be interpreted as a Bayesian treatment that the posterior distribution of the weight is given by a Gaussian distribution \ie~ $q_{\phi}(m_{r,c})=\mathcal{N}(\theta_{r,c}, \sigma^{2}\theta_{r,c}^{2})$. 
The HyperDrop sampling procedure Eq~\eqref{equation:regularization_origin} can be interpreted as rising from a reparameterization of the posterior on the parameter $\Mat{M}$ as shown in Eq~\eqref{equation:regularization_bayesian}. 

\subsection{Residual Hyperbolic Graph Convolutional Network} We then investigate our architecture of $\mathcal{R}$-HGCN and its effect of preventing over-smoothing. In fact, we always combine $\mathcal{R}$-HGCN with initial input. Note that our model is of the form $\mathbb{L}=(\mathcal{L}^d)_{o_1}\times(\mathcal{L}^d)_{o_2}\times\cdots\times(\mathcal{L}^d)_{o_k}$, where $(\mathcal{L}^d)_{o_i}$ are copies of $d$-dimensional Lorentz
spaces with random prescribed origin points $o_i\in \mathcal{L}^d$. Then for any $x=(x_1,\dots,x_k)\in\mathbb{L}$, its Dirichlet energy is defined as 
\begin{equation}\label{eq-def of multiple Dirichlet energy}
	E(x)=\max_{1\leq i\leq k}(\log_{o_i}(x_i)^T\tilde{\Delta}\log_{o_i}(x_i)):=\max_{1\leq i\leq k}E_i(x_i).
\end{equation}
Then by the similar argument with the proof of (\ref{equ:dewoinc}), we can estimate $E_i(H_i^{(l)})$ separately and then opt for the maximal among them, which may be better behaved than any single component due to possible fluctuation.

\paragraph{Network Architecture.}
\label{subsection:Net_archi}
Let $\mathbb{L}=(\mathcal{L}^d)_{o_1}\times(\mathcal{L}^d)_{o_2}\times\cdots\times(\mathcal{L}^d)_{o_k}$ denote the product manifold of $k$ Lorentz models where $d_{j}$ is the dimension of the $j$-th Lorentz model. 
The initial node features $\Mat{H}^{(0)}$ on the product manifold of Lorentz models are given by ($\Vec{o} ~ = [\Vec{o}_{1}, \cdots, \Vec{o}_{j}, \cdots, \Vec{o}_{k}]$)
\begin{equation}
\label{equation:obtain_product_hyper}
\begin{aligned}
\Mat{H}^{(0)} = [ \Mat{H}^{1, {(0)}}&,  \cdots, \Mat{H}^{j, {(0)}}, \cdots, \Mat{H}^{k, {(0)}} ],\\
\Mat{H}^{j, {(0)}}_{i} &= \mathrm{exp}_{\Vec{o}}\big([0, \Mat{X}_{i} ]\big), \\
\end{aligned}
\end{equation}
where $\Mat{H}^{j, {(0)}}_{i}\in \mathbb{R}^{(d_{j}+1)}$ denotes the $i$-th row of the node features $\Mat{H}^{j,{(0)}} \in \mathbb{R}^{n \times (d_{j}+1)}$ on the $j$-th Lorentz component. 

The graph convolution on the product manifold of the Lorentz models combined with HyperDrop is realized by instantiating $f(\cdot)$ in Eq~\eqref{equation:hyper_drop_train} as the hyperbolic graph convolution operator $\mathrm{hgc}(\cdot)$ , \ie~ Eq~\eqref{equation:single_DeepHGCN_layer} becomes
\begin{equation}
\begin{aligned}
\Mat{H}^{(\ell)} = [ & \Mat{H}^{1,{(\ell)}}, \cdots, \Mat{H}^{j, {(\ell)}}, \cdots, \Mat{H}^{k, {(\ell)}} ],\\ 
\Mat{H}_{i}^{j, (\ell)} &= \xi_{i}^{j} \odot \mathrm{hgc}(\Mat{H}^{j, (\ell-1)}), \\
&~\xi_{i}^{j}  \sim \mathcal{N}(1,\sigma^{2}), \\
\end{aligned}
\label{equation:product_DeepHGCN_layer}
\end{equation}
where $\Mat{H}^{j, {(\ell)}} \in \mathbb{R}^{n \times (d_{j}+1)}$ is the node features on the $j$-th Lorentz component at the $\ell$-th layer. 
$\Mat{H}_{i}^{j, {(\ell)}} \in \mathbb{R}^{(d_{j}+1)}$ is the $i$-th row (\ie,~the $i$-th node) of $\Mat{H}^{j, {(\ell)}}$. 
$\odot$ denotes the Lorentz scalar multiplication defined in Definition~\ref{definition:Lorentz_scalemul}. 
$\xi_{i}^{j}$ is the random multiplicative noise drawn from the Gaussian distribution $\mathcal{N}(1,\sigma^{2})$. 
We set $\sigma=\eta/(1-\eta)$ where $\eta$ denotes the drop rate. 
The node features at the last layer can be used for downstream tasks. 
Taking the node classification task as an example, we map node features to the tangent spaces of the product manifolds, and send tangent representations to a fully-connected layer followed by a softmax for classification. 

\section{Experiments}
\label{section:experiments}
Experiments are performed on the semi-supervised node classification task. 
We first evaluate the performance of $\mathcal{R}$-HGCN under different configurations of models, including various graph convolution layers and different structures of product manifolds.
Then, we compare with several state-of-the-art Euclidean GCNs and HGCNs, showing that $\mathcal{R}$-HGCN achieves competitive results. 
Further, we compare with DropConnect\cite{wan2013regularization}, a related regularization mathod for deep GCNs. 


\begin{table}
	\centering
	\begin{tabular}{ccccc}\cmidrule[\heavyrulewidth]{1-5}
		\multirow{1}{*}{\vspace*{8pt}{Datasets}}&\multicolumn{1}{c}{\textsc{Pubmed}} & \multicolumn{1}{c}{\textsc{Citeseer}} & \multicolumn{1}{c}{\textsc{Cora}} & \multicolumn{1}{c}{\textsc{Airport}} \\ 
		\hline
		{Classes} &  	$3$ 		&$6$ 		&$7$      &$4$    \\
		{Nodes} & 		$19,717$ 	&$3,327$ 	&$2,708$  &$3,188$\\
		{Edges} &  		$44,338$ 	&$4,732$ 	&$5,429$  &$18,631$\\
		{Features} &	$500$ 		&$3,703$ 	&$1,433$  &$4$\\
		\cmidrule[\heavyrulewidth]{1-5}
	\end{tabular}
	\caption{Dataset statistics. }
	\label{table:datasets_statistic} 
\end{table}

\subsection{Datasets and Baselines}
We use four standard commonly-used citation network graph datasets : \textsc{Pubmed}, \textsc{Citeseer}, \textsc{Cora} and \textsc{Airport} \cite{sen2008collective}. 
Dataset statistics are summarized in Table \ref{table:datasets_statistic}. Experiment details see in the supplementary material.


\begin{table*}[!htb]
	\center
	\small 
	\centering
			\setlength{\tabcolsep}{0.8mm}{
				\begin{tabular}{ll|cc|cc|cc|cc|cc}
					\toprule[\heavyrulewidth]
					\multirow{2}{*}{Datasets} & \multirow{2}{*}{Methods} &  \multicolumn{2}{c|}{4 layers}  & \multicolumn{2}{c|}{8 layers} & \multicolumn{2}{c|}{16 layers}  & \multicolumn{2}{c}{32 layers}\\
					&  & \multicolumn{1}{c}{Original} & \multicolumn{1}{c|}{HyperDrop}   & \multicolumn{1}{c}{Original} & \multicolumn{1}{c|}{HyperDrop} & \multicolumn{1}{c}{Original} & \multicolumn{1}{c|}{HyperDrop}   & \multicolumn{1}{c}{Original} & \multicolumn{1}{c}{HyperDrop}\\
					\hline
					\multirow{5}{*}{\textsc{Pubmed}}       
					& GCNII   & \multicolumn{2}{c|}{$79.3\pm0.3 $}  & \multicolumn{2}{c|}{$79.9 \pm0.3 $} &
					\multicolumn{2}{c|}{$79.9\pm1.7$} &
					\multicolumn{2}{c}{$80.0\pm1.9$}\\
					& $\mathcal{P}$-HGCN$_{[2 \times 8]}$  &   $79.1 \pm0.2 $  &   $\textbf{79.8}\pm0.3 $  &   $80.0 \pm0.1 $ &   $\textbf{80.3}\pm0.1 $  
					&   $ 79.1\pm0.2 $
					&  $79.2\pm0.3$
					&   $ 79.2\pm0.3 $
					& \multicolumn{1}{c}{ $79.5\pm0.4$}
					\\
					& $\mathcal{P}$-HGCN$_{[4 \times 4]}$  &   $79.0 \pm0.2 $  &   $79.5 \pm0.2 $    &   $79.8 \pm0.1 $  &   $80.1 \pm0.3 $    
					&   $ 79.9\pm0.3 $
					&  $80.0\pm0.2$
					&   $ 79.8\pm0.4 $
					& \multicolumn{1}{c}{ $\textbf{80.3}\pm0.3$}
					\\
					& $\mathcal{P}$-HGCN$_{[8 \times 2]}$ &   $79.0\pm0.2 $  &   $79.4\pm0.2$    &   $79.7\pm0.2$  &   $79.9\pm0.2$    
					&   $ 80.0\pm0.3 $
					&   $\textbf{80.1}\pm0.3$
					&   $ 80.1\pm0.2 $
					&  \multicolumn{1}{c}{ $80.3\pm0.4$}
					\\
					& $\mathcal{P}$-HGCN$_{[16 \times 1]}$  &   $78.7 \pm0.3 $ &   $79.2 \pm0.3 $   &   $79.4 \pm0.1 $   &   $79.9 \pm0.2$
					&   $ 79.3\pm0.3 $
					&   $79.5\pm0.4$
					&   $ 79.2\pm0.2 $
					&   \multicolumn{1}{c}{$80.1\pm0.4$}
					\\
					\hline
					\multirow{5}{*}{\textsc{Citeseer} }      
					& GCNII  & \multicolumn{2}{c|}{$69.3\pm2$
					}  & \multicolumn{2}{c|}{$70.6\pm1.4$
					} & \multicolumn{2}{c|}{$70.5\pm1.3$}
					& \multicolumn{2}{c}{$70.8\pm1.6$}\\
					& $\mathcal{P}$-HGCN$_{[2 \times 8]}$ &   $71.4 \pm0.2 $ &   $72.0 \pm0.4 $   &   $72.1 \pm0.2 $   &   $72.3 \pm0.3 $    &   $ 71.9 \pm0.5 $
					&   $ 71.9 \pm0.7 $ 
					&   $ 72.1 \pm0.6 $ 
					&   \multicolumn{1}{c}{$ 72.3 \pm0.7 $}\\
					& $\mathcal{P}$-HGCN$_{[4 \times 4]}$  &   $71.4 \pm0.2 $   &   $72.0 \pm0.4 $   &   $71.9 \pm0.3 $   &   $72.2 \pm0.3 $ 
					&   $ 71.5 \pm0.5 $
					&   $ 71.7 \pm1.0 $  
					&   $ 71.4 \pm0.2 $
					&   \multicolumn{1}{c}{$ 71.9 \pm0.7 $}
					\\
					& $\mathcal{P}$-HGCN$_{[8 \times 2]}$
					&  $71.2\pm0.8$
					&   $70.4\pm0.9$
					&  $68.6\pm0.8$
					&  $71.1\pm0.7$
					&   $72.0\pm0.2$
					&   $ 71.9 \pm0.9 $
					&   $72.0\pm0.1$
					&   \multicolumn{1}{c}{$ 72.0\pm0.5 $}
					\\
					& $\mathcal{P}$-HGCN$_{[16 \times 1]}$  &   $71.3 \pm0.3 $   &   $\textbf{72.4} \pm0.3 $  &   $71.9 \pm0.3 $   &   $\textbf{72.5} \pm0.5 $ 
					&   $ 72.2 \pm0.4 $
					&   $ \textbf{72.3} \pm0.4 $
					&   $ 72.3 \pm0.6 $
					&   \multicolumn{1}{c}{$ \textbf{72.5} \pm0.9 $}  
					\\
					\hline
					\multirow{5}{*}{\textsc{Cora}}       
					& GCNII & \multicolumn{2}{c|}{$76.6\pm2.4$
					}  & \multicolumn{2}{c|}{ $79.4\pm1.4$
					} & \multicolumn{2}{c|}{$81.3\pm1.0$
					} & \multicolumn{2}{c}{$81.5\pm1.4$
					}\\
					& $\mathcal{P}$-HGCN$_{[2 \times 8]}$ &   $80.3 \pm0.7 $   &   $\textbf{82.0}\pm0.5 $  &   $81.5 \pm0.2 $ &   $\textbf{82.2} \pm0.4 $  
					&   $ 81.8 \pm0.2 $
					&   $ \textbf{82.1} \pm0.2 $
					&   $ 81.9 \pm0.3 $
					&   \multicolumn{1}{c}{$ 82.1 \pm0.1 $}
					   \\
					& $\mathcal{P}$-HGCN$_{[4 \times 4]}$&   $80.3 \pm0.8 $   &   $81.7 \pm0.5 $  &   $81.3 \pm0.2 $   &   $81.9 \pm0.4 $ 
					&   $ 81.7 \pm0.2 $
					&   $ 81.9 \pm0.2 $
					&   $ 81.6 \pm0.5 $
					&   \multicolumn{1}{c}{$ 82.1 \pm0.7 $}
					  \\
					& $\mathcal{P}$-HGCN$_{[8 \times 2]}$
					&  $77.9\pm0.8$
					&   $80.8\pm0.8$
					&  $80.2\pm0.9$
					&   $81.5\pm0.3$
					
					&   $81.7\pm0.2$
					
					&   $81.8\pm0.2$
					&   $81.9\pm0.5$
					&   \multicolumn{1}{c}{$\textbf{82.3}\pm0.8$}
					\\
					& $\mathcal{P}$-HGCN$_{[16 \times 1]}$  &   $79.4 \pm0.8 $   &   $81.9 \pm0.7 $  &   $80.4 \pm0.4 $   &   ${82.5} \pm0.4 $
					&   $  81.6 \pm0.6 $
					&   $ 81.5 \pm0.5 $ 
					&   $  81.6 \pm0.7 $
					&  \multicolumn{1}{c}{ $ 81.4 \pm0.6 $  }  \\
					\hline
					\multirow{5}{*}{\textsc{Airport} }      
					& GCNII  & \multicolumn{2}{c|}{$88.9 \pm0.5 $}  & \multicolumn{2}{c|}{$89.1 \pm0.3 $} & \multicolumn{2}{c|}{$ 89.2 \pm0.4 $}& \multicolumn{2}{c}{$ 89.6 \pm0.7 $}\\
					& $\mathcal{P}$-HGCN$_{[2 \times 8]}$ 
					&  $88.9\pm0.4$
					&  $\textbf{89.1}\pm0.2$
					&   $89.2\pm1.0$
					&   $89.4\pm0.9$
					&   $89.1\pm0.3$
					&  $89.3\pm0.5$
					&   $89.7\pm0.2$
					& \multicolumn{1}{c}{ $90.0\pm0.4$}
					\\
					& $\mathcal{P}$-HGCN$_{[4 \times 4]}$  
					&  $88.6\pm0.6$
					&  $89.0\pm0.2$
					&   $89.3\pm0.4$
					&   $89.4\pm0.3$
					&   $89.6\pm0.1$
					&  $89.6\pm0.2$
					&   $89.7\pm0.1$
					& \multicolumn{1}{c}{ $\textbf{90.2}\pm0.7$}
					\\
					& $\mathcal{P}$-HGCN$_{[8 \times 2]}$ 
					&  $88.7\pm0.2$
					&  $88.7\pm0.4$
					&   $89.3\pm0.5$
					&   $\textbf{89.6}\pm0.7$
					&   $89.4\pm0.2$
					&  $\textbf{89.8}\pm0.5$
					&   $89.6\pm0.4$
					&  \multicolumn{1}{c}{$89.7\pm0.7$}
					\\
					& $\mathcal{P}$-HGCN$_{[16 \times 1]}$ 
					&  $88.5\pm0.7$
					&  $88.6\pm0.4$
					&   $88.6\pm0.3$
					&   $88.5\pm0.5$
					&   $88.7\pm0.2$
					&  $88.9\pm0.5$
					&   $89.0\pm0.3$
					& \multicolumn{1}{c}{ $89.2\pm0.5$}
					\\
					
					\bottomrule[\heavyrulewidth]
			\end{tabular}}
	\setlength{\tabcolsep}{1.0mm}{\caption{Comparisons on various graph convolution layers and different structures of $16$-dimensional product manifolds w and w/o HyperDrop. We also compare with a related work GCNII using $16$-dimensional embedding space. Mean accuracy (\%) and standard deviation are reported. 
		$\mathcal{P}$-HGCN$_{[d \times m]}$ denotes the {$\mathcal{P}$-HGCN} with a product manifold of $m$ $d$-dimensional Lorentz models. 
	}}
	\label{table:depth-table}
\end{table*}

\subsection{Validation Experiments}
\label{subsection:analytical_experiments}
Here we demonstrate the effectiveness of the $\mathcal{R}$-HGCN and our regularization method under different model configurations. 
For $\mathcal{R}$-HGCN, increasing the number of hyperbolic graph convolution layers almost always brings improvements on three datasets. 

\begin{table*}[!th]  
	\centering
	{
		\begin{tabular}{clccc}\cmidrule[\heavyrulewidth]{1-5}
			& 
			{Methods} & \multicolumn{1}{c}{\textsc{Pubmed}} & \multicolumn{1}{c}{\textsc{Citeseer}}  &\multicolumn{1}{c}{\textsc{Cora}} \\
			\cmidrule{1-5}
			\multirow{6}{*}{\rotatebox{90}{}\rotatebox{90}{\hspace*{-6pt} Euclidean}} 
			& {GCN}\cite{kipf2016semi}   & $79.1\pm0.3$ & $71.2\pm0.6$  &  $81.3\pm0.5$ \\
			& {GAT}\cite{velivckovic2017graph}   & $77.7\pm0.2$ & $70.9\pm0.4$  &  $82.4\pm0.6$ \\
			& {GraphSage}\cite{hamilton2017inductive}   & $77.3\pm0.3$ & $67.8\pm1.1$  &  $77.3\pm0.8$ \\
			& {SGC}\cite{wu2019simplifying} &  $69.3\pm0.0$ & $78.9\pm0.0$  &  $80.9\pm0.0$ \\
			& {APPNP}\cite{klicpera2018predict} &   $80.1\pm0.2$ & $71.6\pm0.4$  &  $\textbf{83.7}\pm0.5$ \\
			& {GCNII}(8)\cite{chen2020simple} &   $79.9\pm0.3$ & $72.4\pm0.9$  &  $83.5\pm0.7$ \\
			\cmidrule{1-5}
			\multirow{2}{*}{\rotatebox{90}{}\rotatebox{90}{\hspace*{-50pt} Hyperbolic}}
			& {HGCN} \cite{chami2019hyperbolic} &   $\textbf{80.3}\pm0.3$ & - &  $79.9\pm0.2$ \\ 
			& {H2HGCN}\cite{dai2021hyperbolic} &   $79.9\pm0.5$ & -  &  $82.8\pm0.4$ \\
			& {$\kappa$ GCN} \cite{bachmann2020constant} &   $78.3\pm0.6$ & $70.7\pm0.5$  &  $80.0\pm0.6$ \\
			& {LGCN} \cite{zhang2021lorentzian} &   $78.6\pm0.7$ & $71.9\pm0.7$  &  $83.3\pm0.7$ \\
			& $\mathcal{P}$-HGCN$_{[16 \times 1]}$(8) &   $79.4\pm0.1$ & $71.9\pm0.3$  &  $80.4\pm0.4$ \\
			& $\mathcal{P}$-HGCN$_{[16 \times 1]}$(8)+HyperDrop &   $79.9\pm0.2$ & $\textbf{72.5}\pm0.5$  &  $82.5\pm0.4$ \\
			& $\mathcal{P}$-HGCN$_{[2 \times 8]}$(8) &   $80.0\pm0.1$ & $72.1\pm0.2$  &  $81.5\pm0.2$ \\
			& $\mathcal{P}$-HGCN$_{[2 \times 8]}$(8)+HyperDrop &   $\textbf{80.3}\pm0.1$ & $72.3\pm0.3$  &  $82.2\pm0.4$ \\
			\cmidrule[\heavyrulewidth]{1-5}
	\end{tabular} }
\caption{Mean accuracy (\%) and standard deviation on \textsc{Pubmed}, \textsc{Citeseer}, and \textsc{Cora}.
	We set the dimensions of embedding spaces to $16$ for all methods and the number of graph convolution layers to $8$ (number in parentheses) for deep models, \ie, GCNII and $\mathcal{P}$-HGCN. 
	$\mathcal{P}$-HGCN$_{[d \times m]}$ denotes the $\mathcal{P}$-HGCN with a product manifold of $m$ $d$-dimensional Lorentz models. 
}
\label{table:table_sort}
\end{table*}
\begin{table*}[htbp]
	\centering
	\setlength{\tabcolsep}{1.mm}
	\begin{tabular}{c|cc|cc|cc|cc}
		\hline
		\multirow{2}{*}{Layers} & \multicolumn{2}{c|}{$\mathcal{P}$-HGCN$_{[2 \times 8]}$} & \multicolumn{2}{c|}{$\mathcal{P}$-HGCN$_{[4 \times 4]}$} &\multicolumn{2}{c|}{$\mathcal{P}$-HGCN$_{[8 \times 2]}$} & \multicolumn{2}{c}{$\mathcal{P}$-HGCN$_{[16 \times 1]}$} \\ 
		& with IRC & w/o IRC & with IRC & w/o IRC & with IRC & w/o IRC & with IRC & w/o IRC \\
		\hline
		2  & $78.9\pm0.4$ & $78.9\pm0.2$ & $79.0\pm0.3$ & $78.8\pm0.3$ & $78.8\pm0.3$ & $78.9\pm0.4$ & $78.7\pm0.2$ & $78.6\pm0.4$ \\
		4  & $79.1\pm0.4$ & $79.3\pm0.2$ & $79.0\pm0.2$ & $79.0\pm0.6$ & $79.2\pm0.3$ & $78.7\pm0.4$ & $79.2\pm0.3$ & $78.6\pm0.4$ \\
		8  & $80.3\pm0.0$ & $78.6\pm0.2$ & $80.2\pm0.2$ & $79.1\pm0.2$ & $80.1\pm0.1$ & $78.5\pm0.6$ & $79.8\pm0.2$ & $76.1\pm3.5$ \\
		16 & $79.2\pm0.3$ & $29.8\pm11.1$ & $80.0\pm0.2$ & $60.1\pm7.8$ & $80.1\pm0.3$ & $60.1\pm7.8$ & $79.5\pm0.4$ & $49.9\pm2.1$ \\
		\hline
	\end{tabular}
	\caption{ Testing accuracy (\%) comparisons on different layers and model structures w and w/o hyperbolic residual connection(HRC).
		$\mathcal{P}$-HGCN$_{[d \times m]}$ denotes the $\mathcal{P}$-HGCN with a product manifold of $m$ $d$-dimensional Lorentz models. }
	\label{table:example}
\end{table*}
\begin{table}[htb]
	\small
	\centering
	\begin{tabular}{cccc}\cmidrule[\heavyrulewidth]{1-4}
		\multirow{1}{*}{\vspace*{8pt}{Methods}}&\multicolumn{1}{c}{\textsc{Pubmed}} & \multicolumn{1}{c}{\textsc{Citeseer}} & \multicolumn{1}{c}{\textsc{Cora}}  
		\\ 
		\hline
		w/o dropout&  $79.4\pm0.1$  &  $71.9\pm0.3$  &  $80.3\pm0.4$  \\
		DropConnect &  $79.7\pm0.3$  &  $71.9\pm0.3$  &  $83.0\pm0.6$  \\
		HyperDrop &  $79.9\pm0.2$  &  $72.5\pm0.5$  &  $82.5\pm0.4$  \\
		Both &  $79.9\pm0.3$  &  $72.7\pm0.5$  &  $83.5\pm0.9$  \\
		\cmidrule[\heavyrulewidth]{1-4}
	\end{tabular}
	\caption{Comparisons of HyperDrop and DropConnect. 
	}
	\label{table:hyperdrop_dropconnect} 
\end{table}

The criterion for judging the effectiveness of HyperDrop is to test whether the performance of $\mathcal{R}$-HGCN is improved with the help of HyperDrop. 
In Table~\ref{table:depth-table}, we report the performance of HyperDrop with various graph convolution layers and structures of product manifolds on \textsc{Pubmed}, \textsc{Citeseer} \textsc{Cora} and \textsc{Airport}. 
We observe that in most experiment, HyperDrop improves the performance of $\mathcal{R}$-HGCN. 
For example, on \textsc{Cora}, $\mathcal{R}$-HGCN$_{[2 \times 8]}$ obtains $1.7\%$, $0.7\%$, $0.3\%$ and $0.2\%$ gains with $4$, $8$, $16$ and $32$ layers.
The stable improvements demonstrate that HyperDrop can effectively improve the generalization ability of $\mathcal{R}$-HGCN.

We also compare $\mathcal{R}$-HGCN with a deep Euclidean method, GCNII. 
The hyperbolic residual connection and hyperbolic identity mapping in $\mathcal{R}$-HGCN are inspired by GCNII. 
The main difference between $\mathcal{R}$-HGCN and GCNII is, $\mathcal{R}$-HGCN performs graph representation learning in hyperbolic spaces while GCNII is in Euclidean spaces. 
As shown in Table~\ref{table:depth-table}, $\mathcal{R}$-HGCN shows superiority compared to GCNII. 
Actually, $\mathcal{R}$-HGCN is only baseline model we developed for evaluating the effectiveness of HyperDrop, and we give up extra training tricks for clear evaluations. 
For example, in Section~\ref{subsubsection:HyperDrop_DropConnect}, $\mathcal{R}$-HGCN obtains the same mean accuracy $83.5\%$ as GCNII with $8$ layers on \textsc{Cora} while using HyperDrop and DropConnect\cite{wan2013regularization} together. 
DropConnect is used in other hyperbolic graph convolutional networks, such HGCN \cite{chami2019hyperbolic} and LGCN \cite{zhang2021lorentzian}. 
We claim that the superior performance of $\mathcal{R}$-HGCN compared to GCNII benefits from the representing capabilities of hyperbolic spaces while dealing with hierarchical-structure data. 
It confirms the significance of hyperbolic representation learning. 
\subsection{Ablation Experiments}
We conducted ablation experiments on the \textsc{Pubmed} dataset to observe the effect of our proposed  hyperbolic residual connection on $\mathcal{R}$-HGCN performance in different dimension selection approaches, respectively. As can be seen from Tabel~\ref{table:example}, without the hyperbolic residual connection, the fortunate performance of the model shows a different degree of decrease respectively. Moreover, as the number of model layers increases, the decline in model performance is more pronounced. The experimental results prove that hyperbolic residual connection has a great helpful effect on the model performance.

\subsection{Performance Comparisons}
The comparisons with several state-of-the-art Euclidean GCNs and HGCNs are shown in Table~\ref{table:table_sort}. 
We have three observations. 
First and most importantly, compared with other HGCNs that are typically shallow models, $\mathcal{R}$-HGCN shows better results on \textsc{Pubmed} and \textsc{Citeseer}. 
Through, on \textsc{Pubmed}, {HGCN} also achieves the best accuracy, $80.3\%$.
Note that {HGCN} uses extra link prediction task as pre-training model while $\mathcal{R}$-HGCN does not use this training trick for a clear evaluation of HyperDrop; and the performance of HGCN decreases when the link prediction pre-training is not used. 
On \textsc{Cora}, {LGCN} achieves the highest mean accuracy $83.3\%$ among HGCNs. 
Note that both HGCN and LGCN utilize DropConnect \cite{wan2013regularization} technique for training. 
As shown in Section~\ref{subsubsection:HyperDrop_DropConnect}, 
$\mathcal{R}$-HGCN obtains $83.5\%$ mean accuracy on \textsc{Cora} while also using DropConnect, that is $0.2\%$ higher than that of {LGCN}. 
Second, both $\mathcal{R}$-HGCN$_{[16 \times 1]}$ and $\mathcal{R}$-HGCN$_{[2 \times 8]}$ benefit from HyperDrop on three datasets. 
It proves HyperDrop alleviates over-fitting issue in hyperbolic graph convolutional network and improves the generalization ability of $\mathcal{R}$-HGCN on the test set. 
Third, compared with Euclidean GCNs, $\mathcal{R}$-HGCN combined with HyperDrop achieves the best results on \textsc{Pubmed} and \textsc{Citeseer}. 
It confirms the superiority of hyperbolic representation learning while modeling graph data. 


\subsection{Comparisons with DropConnect}
\label{subsubsection:HyperDrop_DropConnect}
Table~\ref{table:hyperdrop_dropconnect} shows the performance of HyperDrop and DropConnect\cite{wan2013regularization} on \textsc{pubmed}, \textsc{Citeseer}, and \textsc{Cora} using a $16$-dimensional $\mathcal{R}$-HGCN. 
Since there is no dropout method tailed for hyperbolic representations before HyperDrop, some works \cite{chami2019hyperbolic,zhang2021lorentzian} use DropConnect as a regularization. 
DropConnect is one of variants of dropout methods that randomly zeros out elements of the Euclidean parameters in model, and it can be used in hyperbolic graph convolutional network as the parameters in hyperbolic graph convolutional network are Euclidean.

For DropConnect, we search the drop rate from $0.1$ to $0.9$ and report the best results. 
DropConnect obtains improvements on \textsc{Pubmed} and \textsc{Cora} but not on \textsc{Citeseer}. 
In contrast, HyperDrop achieves stable improvements on three datasets, and higher mean accuracy on \textsc{Pubmed} and \textsc{Citeseer} compared to DropConnect. 
HyperDrop and DropConnect are two dropout methods that the former works on hyperbolic representations and the latter works on Euclidean parameters. 
They can work together effectively for a better generalization of $\mathcal{R}$-HGCN. 
As the results on \textsc{Citeseer} and \textsc{Cora} show, using HyperDrop and DropConnect together has better performance than using only HyperDrop or DropConnect individually.
\section{Conclusion}

In this paper, we have proposed $\mathcal{R}$-HGCN, a product manifold based residual hyperbolic graph convolutional network for overcoming the over-smoothing problem. 
The residual connections can prevent node representations from being indistinguishable by hyperbolic residual connection and hyperbolic identity mapping. 
The product manifold with different origin points also provides a wider range of perspectives of data. 
A novel hyperbolic dropout method, HyperDrop, is proposed to alleviate the over-fitting issue while deepening models. 
Experiments have demonstrated the effectiveness of $\mathcal{R}$-HGCNs  under various graph convolution layers and different structures of product manifolds. 
\section*{Acknowledgements}
This work was supported by the Natural Science Foundation of China (NSFC) under Grants No. 62172041 and No. 62176021, Natural Science Foundation of Shenzhen under Grant No. JCYJ20230807142703006, and 2023 Shenzhen National Science Foundation(No. 20231128220938001).

\bibliographystyle{plainnat}
\bibliography{egbib.bib}

\end{document}